\pdfoutput=1

\documentclass[11pt]{article}

\usepackage[final]{acl}
\usepackage{times}
\usepackage{latexsym}
\usepackage[T1]{fontenc}
\usepackage[utf8]{inputenc}
\usepackage{microtype}
\usepackage{inconsolata}
\usepackage{graphicx}
\usepackage{hyperref}
\usepackage{amsmath}
\usepackage{amssymb}
\usepackage{mathtools}
\usepackage{amsthm}
\usepackage{bm}
\usepackage{thmtools}
\usepackage{thm-restate}
\usepackage{multirow,adjustbox,makecell,arydshln} 
\usepackage{diagbox}
\usepackage{longtable}
\usepackage{color, colortbl}
\usepackage{booktabs}
\usepackage{xcolor}
\theoremstyle{plain}
\newtheorem{theorem}{Theorem}

\newtheorem*{theoremr*}{Theorem~\ref{th1}}
\newtheorem*{theoremr2*}{Theorem~\ref{th2}}
\newtheorem*{corr*}{Corollary~\ref{cor1}}
\newtheorem*{propositionr*}{Proposition~\ref{prop1}}
\newtheorem*{propositionr2*}{Proposition~\ref{p2}}
\theoremstyle{definition}
\newtheorem{definition}[theorem]{Definition}

\theoremstyle{remark}

\usepackage{amsfonts}
\usepackage{enumitem} 
\usepackage{tcolorbox} 

\renewcommand{\vec}[1]{\boldsymbol{\mathbf{#1}}}

\DeclareMathOperator{\KL}{KL}
\DeclareMathOperator{\RL}{RL}
\DeclareMathOperator{\OT}{OT}

\DeclareMathOperator{\relu}{ReLU}
\DeclareMathOperator{\linear}{Linear}
\DeclareMathOperator{\drop}{Dropout}

\DeclareMathOperator{\SSW}{SSW}

\DeclareMathOperator{\vmf}{vMF}
\DeclareMathOperator{\mvmf}{MvMF}

\usepackage{tcolorbox}
\tcbuselibrary{listingsutf8}
\newtcbox{\highlightpair}{on line, box align=base, colframe=gray!40, colback=gray!40, sharp corners, boxrule=0pt, arc=1pt, boxsep=0pt, left=1pt, right=1pt, top=0.5pt, bottom=0.5pt}

\renewcommand{\vec}[1]{\boldsymbol{\mathbf{#1}}}

\newcommand{\cmmnt}[1]{}
\title{S2WTM: Spherical Sliced-Wasserstein Autoencoder for Topic Modeling}

\author{Suman Adhya \and Debarshi Kumar Sanyal \\
    Indian Association for the Cultivation of Science\\
         \texttt{\href{mailto:adhyasuman30@gmail.com}{adhyasuman30@gmail.com}, \href{mailto:debarshi.sanyal@iacs.res.in}{debarshi.sanyal@iacs.res.in}}
}

\begin{document}
\maketitle
\begin{abstract} 
Modeling latent representations in a hyperspherical space has proven effective for capturing directional similarities in high-dimensional text data, benefiting topic modeling. Variational autoencoder-based neural topic models (VAE-NTMs) commonly adopt the von Mises-Fisher prior to encode hyperspherical structure. However, VAE-NTMs often suffer from posterior collapse, where the KL divergence term in the objective function highly diminishes, leading to ineffective latent representations. To mitigate this issue while modeling hyperspherical structure in the latent space, we propose the Spherical Sliced Wasserstein Autoencoder for Topic Modeling (S2WTM). S2WTM employs a prior distribution supported on the unit hypersphere and leverages the Spherical Sliced-Wasserstein distance to align the aggregated posterior distribution with the prior. Experimental results demonstrate that S2WTM outperforms state-of-the-art topic models, generating more coherent and diverse topics while improving performance on downstream tasks.

\noindent
\includegraphics[width=1.25em,height=1.25em]{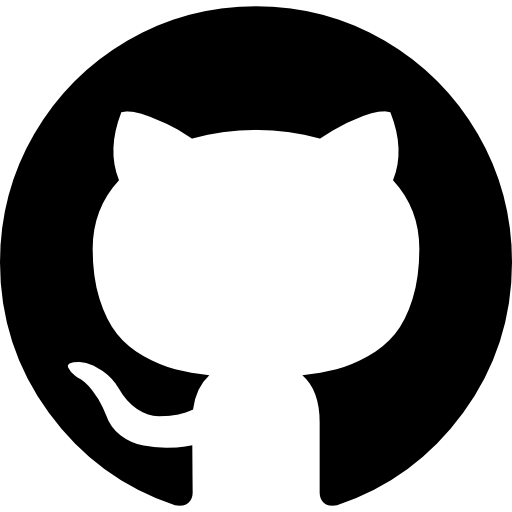}\hspace{0.5em}
\href{https://github.com/AdhyaSuman/S2WTM}{\texttt{github.com/AdhyaSuman/S2WTM}}
\end{abstract}

\section{Introduction}
Topic modeling aims to uncover latent topics in a document corpus by representing each document as a probabilistic mixture of topics. Latent Dirichlet Allocation (LDA) \cite{blei2003latent} and Probabilistic Latent Semantic Indexing (pLSI) \cite{hofmann2013probabilistic} typically assume a multinomial distribution over words. However, this assumption treats words as independent, ignoring important syntactic and semantic relationships. Moreover, classical topic models rely on computationally expensive inference techniques, such as Gibbs sampling and variational inference.

To address these limitations, Variational Autoencoder-based Neural Topic Models (VAE-NTMs) were introduced by \citet{miao2016neural}, offering a more scalable and flexible alternative. VAE-NTMs typically use a Gaussian prior for the latent space. However, in high-dimensional settings, Gaussian distributions exhibit the ``soap bubble effect" \cite{davidson2018hyperspherical}, where probability mass concentrates near the surface of a hypersphere rather than around the mean. This effect undermines the utility of Euclidean distances, which tend to become uniform and less informative as dimensionality increases—a core aspect of the curse of dimensionality \cite{aggarwal2001surprising}. In contrast, cosine similarity, which measures the angular difference between vectors, remains a more robust metric in such regimes \cite{dhillon2003modeling}. These observations motivate the use of spherical latent space modeling, where data is embedded on the surface of a hypersphere and similarities are measured via angles rather than distances. This is particularly effective in large-scale applications involving directional data, such as text categorization and gene expression analysis \cite{dhillon2003modeling, banerjee2005clustering}, where the orientation of data vectors carries meaningful information. The von Mises-Fisher (vMF) distribution naturally models such data and has demonstrated superior performance over models with Gaussian priors in clustering and topic modeling tasks \cite{reisinger2010spherical, li2016integrating, batmanghelich2016nonparametric}. Recent advancements in topic modeling, such as vONTSS \cite{xu2023vontss}, have leveraged the $\mathcal{S}$-VAE framework and integrated supervised signals via optimal transport to enhance topic coherence and classification performance.

Despite these advancements, VAE-based models often suffer from \textit{posterior collapse} \cite{bowman2016generating_high}, wherein the latent representations fail to encode meaningful information. This issue occurs when the KL divergence term in the loss function approaches zero, causing the posterior distribution to converge to the uninformative prior. A detailed analysis of this phenomenon, including its formal definition and decomposition via ELBO surgery, is provided in Appendix~\ref{sec:ap_elbo_surgery}.

To address these challenges, we propose S2WTM (Spherical Sliced-Wasserstein Autoencoder for Topic Modeling), an autoencoder-based topic model that models the latent space on the unit hypersphere. S2WTM introduces a flexible prior design by treating the choice of prior distribution as a hyperparameter, supporting three alternatives: the vMF distribution, a Mixture of vMF (MvMF) distributions, and the uniform distribution on the hypersphere. To align the aggregated posterior with the chosen prior while preserving spherical geometry, S2WTM employs the Spherical Sliced-Wasserstein (SSW) distance, effectively mitigating the posterior collapse problem.

\paragraph{Contributions:} In summary, our work presents the following key contributions:
\begin{itemize}
    \item We introduce S2WTM, a neural topic model leveraging the Spherical Sliced-Wasserstein autoencoder framework to align the aggregated posterior with a hyperspherical prior, ensuring effective latent space modeling.

    \item We explore three distinct choices of prior distributions defined on the unit hypersphere for modeling the latent space.
    
    \item We evaluate the performance of S2WTM using quantitative metrics, qualitative analysis, task-specific performance, and LLM-based human-comparable evaluations, demonstrating its advantages over state-of-the-art models.
    
    \item We empirically validate the benefits of moving beyond the conventional Euclidean latent space, highlighting performance improvements in topic modeling.
\end{itemize}

\section{Preliminaries}  
This section introduces three selected distributions defined on a unit hypersphere. We then discuss optimal transport metrics, focusing on the Wasserstein distance, sliced Wasserstein distance, and spherical sliced-Wasserstein distance.

\subsection{Distributions on the Unit Hypersphere}
\paragraph{vMF distribution:} The vMF distribution is defined on the unit $(K-1)$-sphere $\mathcal{S}^{K-1}$ embedded in $\mathbb{R}^K$. The vMF distribution is parameterized by its mean direction ($\vec{\mu} \in \mathcal{S}^{K-1}$), a unit vector indicating the central direction of the distribution, and its concentration ($\kappa \in \mathbb{R}^+ \cup \{0\} $), a non-negative scalar controlling the dispersion around $\vec{\mu}$. A larger $\kappa$ indicates a higher concentration. For $\vec{x} \in \mathcal{S}^{K-1}$ with $\kappa \geq 0$ the vMF density function is: $\vmf(\vec{x} ; \vec{\mu}, \kappa) = c_p(\kappa) \exp{(\kappa \vec{\mu} ^\intercal \vec{x})}$ where $c_K(\kappa)$ is the normalization constant: $c_K(\kappa) = \frac{\kappa^{\frac{K}{2}-1}}{(2\pi)^\frac{K}{2} I_{\frac{K}{2}-1}(\kappa)}$ with $I_v$ denotes the modified Bessel function of the first kind of order $v$. The detailed sampling from the vMF distribution \cite{ulrich1984computer} is described in Appendix \ref{sec:ap_prior}.

\paragraph{MvMF distribution:} The MvMF distribution is a convex combination of vMF distributions. If the mixture contains $T$ vMF components, then the density is defined for all points $\vec{x} \in \mathcal{S}^{K-1}$ as $\mvmf(\vec{x};\vec{\Psi}) = \sum_{t=1}^{T} \alpha_t f_t(\vec{x}|\vec{\psi}_t)$. Here, each component $f_t$ is a vMF distribution with parameters $\vec{\psi}_t = (\vec{\mu}_t, \kappa_t)$, and $\vec{\Psi}$ represents the set of all such parameters. The mixture proportions $(\alpha_t)_{1 \leq t \leq T}$ satisfy $\alpha_t \geq 0$ and $\sum_{t=1}^{T} \alpha_t = 1$. The sampling technique from MvMF requires first selecting a component $i \sim \operatorname{Cat}(\alpha_1, \alpha_2, ..., \alpha_T)$, then generating a sample from the chosen vMF distribution.

\paragraph{Uniform distribution on the unit hypersphere:} The uniform distribution on $\mathcal{S}^{K-1}$ assigns equal probability density to all directions. Sampling is performed by drawing from a standard Gaussian in $\mathbb{R}^K$ and applying $\ell_2$ normalization. 

\subsection{Spherical Sliced-Wasserstein Distance:}
The Spherical Sliced-Wasserstein (SSW) distance is a variant of the Sliced-Wasserstein (SW) distance, a computationally efficient approximation of the Wasserstein distance. SSW adapts SW for distributions on the unit hypersphere or other spherical domains.  We first review the Wasserstein and SW distances.

\begin{definition}[Wasserstein Distance \cite{villani2009optimal, bianchini2011optimal}]
Let $p \geq 1$. The $p$-Wasserstein distance between $\mu,\nu\in\mathcal{P}_p(\mathbb{R}^d)$ is
\begin{equation}
    W_p^p(\mu,\nu) = \inf_{\gamma\in\Gamma(\mu,\nu)}\ \int_{\mathbb{R}^d \times \mathbb{R}^d} d(x,y)^p\ \mathrm{d}\gamma(x,y)
\end{equation}
where $\mathcal{P}_p(\mathbb{R}^d)$  is the set of Borel probability measures on $\mathbb{R}^d$ with finite $p$-moments, and $\Gamma(\mu,\nu)$ are joint measures with marginals $\mu$ and $\nu$.
\end{definition}

Computing $W_p$ is computationally expensive, scaling as $\mathcal{O}(n^3\log n)$ for discrete measures with $n$ samples \cite{peyre2019computational}. However, $W_p$ has a closed form for univariate distributions:
\begin{equation} \label{eq:1D_Wp}
    W_p ^p (\mu, \nu) = \int_0 ^1 |F_\mu^{-1}(t) - F_\nu^{-1}(t)|^p dt,
\end{equation}
where $F_\mu^{-1}$ and $F_\nu^{-1}$ are the quantile functions. This leads to the SW, exploiting the efficiency of one-dimensional optimal transport.

\begin{definition}[Sliced Wasserstein Distance \cite{rabin2012wasserstein, bonneel2015sliced}]
For $p \geq 1$, the $p$-SW between $\mu,\nu\in\mathcal{P}_p(\mathbb{R}^d)$ is
\begin{equation}
    SW_p^p (\mu, \nu) = \int_{S^{d-1}} W_p^p\Big((\mathcal{R}\mu)^\theta, (\mathcal{R}\nu)^\theta\Big) d\theta
\end{equation}
where $(\mathcal{R}\eta)^\theta$ is the Radon transform, projecting $\eta$ onto the line spanned by $\theta \in \mathcal{S}^{d-1}$.
\end{definition}

SW reduces computation to multiple one-dimensional problems. However, it requires many projections, especially in high dimensions.  Critically, SW is not well-suited for spherical data. The SSW distance addresses this by incorporating spherical geometry.

\begin{definition}[Spherical Sliced Wasserstein Distance \cite{bonet2023spherical}]
For $p \geq 1$, the $p$-SSW between $\mu,\nu\in\mathcal{P}_p(\mathbb{R}^d)$ is
\begin{equation} \label{eq:SSW}
    \SSW_p^p (\mu, \nu) = \int_{S^{d-1}} W_p^p\Big((\tilde{\mathcal{R}}\mu)^\theta, (\tilde{\mathcal{R}}\nu)^\theta\Big) d\theta
\end{equation}
where $(\tilde{\mathcal{R}}\eta)^\theta$ is the spherical Radon transform, mapping $\eta$ to a substructure of the sphere (e.g., a great circle).
\end{definition}

\section{Related Work}
\citet{reisinger2010spherical} pioneered the idea of spherical latent space modeling in topic modeling by introducing the Spherical Admixture Model (SAM). SAM replaces the multinomial likelihood in LDA with the vMF distribution, representing documents as unit vectors on the hypersphere. This approach captures directional semantic relationships more effectively than traditional Euclidean representations and lays the groundwork for subsequent research in spherical topic models.

\citet{li2016integrating} advanced this line of work by proposing the mix-vMF topic model (MvTM) that integrates word embeddings using MvMF distributions, enabling the representation of complex topic structures and improving topic coherence.

\citet{batmanghelich2016nonparametric} further extended spherical topic modeling by introducing a nonparametric model that incorporates word embeddings via the vMF distribution and employs a Hierarchical Dirichlet Process (HDP) to automatically infer the number of topics.

More recently, \citet{ennajari2022embedded} proposed the Embedded Spherical Topic Model (ESTM) for supervised learning, which integrates semantic knowledge from word and knowledge graph embeddings within a hyperspherical latent space. ESTM improves both topic interpretability and predictive performance by modeling documents and topics on the sphere and leveraging variational inference techniques tailored to the vMF distribution.

In the neural setting, \citet{xu2023vontss} introduced vONT, an unsupervised topic model based on the $\mathcal{S}$-VAE framework, and vONTSS, its semi-supervised extension. While vONT models topics using latent variables constrained to the unit hypersphere, vONTSS further aligns latent representations with supervision signals using an optimal transport loss, thereby enhancing topic relevance and classification accuracy. However, like any VAE-based model, its reliance on KL divergence to regulate the posterior distribution to the prior distribution can lead to the posterior collapse problem \cite{bahuleyan2019stochastic}, which may hinder the expressiveness of the latent space if not carefully mitigated.

In contrast, S2WTM addresses these limitations by replacing KL divergence with the Spherical Sliced-Wasserstein (SSW) distance, promoting alignment between the aggregated posterior and hyperspherical priors. This approach preserves spherical geometry and mitigates posterior collapse, distinguishing S2WTM from both classical and neural spherical topic models.

\begin{figure*}
    \centering
    \includegraphics[width=.7\linewidth]{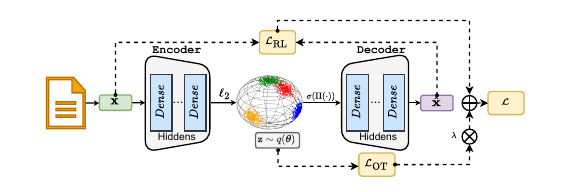}
    \caption{Proposed framework for S2WTM.}
    \label{fig:framework}
\end{figure*}

\section{Proposed Methodology}
The proposed S2WTM model is based on the Wasserstein Autoencoder framework \cite{tolstikhin2018wasserstein}, employing an encoder-decoder architecture. The framework is illustrated in Fig. \ref{fig:framework}, and a detailed description of the methodology is provided below.

\subsection{Choice of Prior Distribution}  
To encourage a hyperspherical structure in the latent space, rather than assuming a standard Euclidean geometry, S2WTM regulates the encoder-generated posterior distribution to align with a prior distribution $ p_{\vec{\theta}} $ defined over the unit hypersphere $ \mathcal{S}^{K-1} $. We consider three types of priors: (i) the vMF distribution, $ p_{\vec{\theta}} \sim \vmf(\vec{\mu}, \kappa) $; (ii) the Mixture of vMF (MvMF) distributions, $ p_{\vec{\theta}} \sim \mvmf(\vec{\Psi}) $; and (iii) the uniform distribution on the hypersphere, $ p_{\vec{\theta}} \sim \mathcal{U}(\mathcal{S}^{K-1})$.  These priors offer varying inductive biases, enabling S2WTM to model diverse latent topic structures while adapting to the characteristics of different datasets. The alignment between the aggregated posterior and the chosen prior is enforced using the SSW distance.

\subsection{Encoder Network}
The encoder processes a document representation, denoted as $\vec{x} \in \mathbb{R}^V$ (where $V$ represents the vocabulary size), and maps it to a low-dimensional representation $\vec{z} \in \mathcal{S}^{K-1}$, with $K$ being the number of topics to be learned. The transformation sequence is as follows: $\big[\linear(V, H') \rightarrow \drop(p) \rightarrow \relu \rightarrow  \linear(H', H'') \rightarrow \drop(p) \rightarrow \relu \rightarrow \linear(H'', K) \rightarrow \operatorname{L2Norm}(\cdot) \big]$. Here, $\text{Linear}(M, N)$ maps an input vector of size $M$ to $N$, $\text{Drop}(p)$ applies dropout with probability $p$, $\relu$ is an activation function defined as $\relu(x) = \max(0, x)$, and $\operatorname{L2Norm}$ normalizes the output onto the unit hypersphere $\mathcal{S}^{K-1}$.

S2WTM employs a deterministic encoder, which learns the \textit{aggregate posterior} distribution: $q(\vec{\theta}) = \int q(\vec{\theta|\vec{x})} p(\vec{x})\: d\vec{x}$ where $p(\vec{x})$ denotes the the empirical data distribution. The encoder generates $\vec{z} \sim q(\vec{\theta})$ such that $\vec{z} \in \mathcal{S}^{K-1}$.

\subsection{Decoder Network}
Given the encoder's generated latent representation $\vec{z}$, it is transformed into the document-topic distribution vector through a projection network followed by $\operatorname{Softmax}$ operation. Finally, the decoder maps it to $\hat{\vec{x}} \in \mathbb{R}$, representing a probability distribution over vocabulary words. The transformation follows: $\big[\operatorname{Linear}(K, H) \rightarrow \operatorname{Dropout}(p) \rightarrow \operatorname{ReLU} \rightarrow \operatorname{Linear}(H, V) \rightarrow \operatorname{Softmax}(\cdot) \big]$. Here, $\operatorname{Softmax}$ is applied to the output to generate the document-word distribution $\hat{\vec{x}}$.

The decoder's weight matrix defines the \textit{topic-word matrix}, $\beta \in \mathbb{R}^{K \times V}$. Following \citet{srivastava2017autoencoding}, we relax the simplex constraint on $\beta$, which is an empirically validated approach that improves the quality of topics.

\subsection{Training Objective}
S2WTM optimizes an objective comprising a reconstruction loss and a regularization term. Unlike VAEs, where regularization is applied to the posterior distribution $q(\vec{\vec{\theta}|\vec{x}})$, S2WTM regularizes the aggregated posterior $q(\vec{\theta})$ using the SSW. In VAEs, ``ELBO surgery" \cite{hoffman2016elbo} decomposes the regularization term into two non-negative components: (1) the mutual information between the input variable $\vec{x}$ and its latent representation $\vec{\theta}$, and (2) the KL divergence between the prior $p(\vec{\theta})$ and the aggregated posterior $q(\vec{\theta})$. Minimizing the KL term reduces mutual information, often leading to posterior collapse (uninformative latent representations) (see Appendix \ref{sec:ap_elbo_surgery}). To mitigate this, S2WTM replaces KL regularization with SSW, ensuring effective alignment between the aggregated posterior and the prior while preserving hyperspherical geometry. The objective function for S2WTM is defined as:
\begin{align}
    \mathcal{L} &= \mathcal{L}_{\RL} + \lambda \mathcal{L}_{\OT} \\ \nonumber
                &= \inf_{q(\vec{\theta}|\vec{x})} \mathbb{E}_{p(\vec{x})} \mathbb{E}_{q(\vec{\theta}|\vec{x})} [c(\vec{x}, \hat{\vec{x}})] + \lambda \SSW_2 ^2 (q_{\vec{\theta}}, p_{\vec{\theta}}).
\end{align} 
Here $p(\vec{x})$ is the data distribution. The reconstruction loss $c(\vec{x}, \hat{\vec{x}})$, is calculated as the cross-entropy between the input document representation $\vec{x}$ and decoder generated document-word distribution $\hat{\vec{x}}$: $\mathcal{L}_{\RL} = -\sum_{i=1}^V x_i \log{\hat{x}_i}$ where $V$ is the vocabulary size. The regularization term employs $\SSW_2 ^2$ defined as:
\begin{equation} \label{eq:L_SSW}
    \SSW_2 ^2 (q_{\vec{\theta}}, p_{\vec{\theta}}) \approx \frac{1}{M} \sum_{i=1}^M  W_2^2\Big((\tilde{\mathcal{R}}_i q_{\vec{\theta}}), (\tilde{\mathcal{R}}_ip_{\vec{\theta}})\Big)
\end{equation}  
where $M$ is the number of random projections, $\tilde{\mathcal{R}}_i$ denotes the $i^{th}$ spherical radon transform. With an increasing number of projections, the SSW distance more accurately captures the structure of the data distribution. In practice, the reconstruction loss often significantly outweighs the regularization term $\SSW_2^2 (q_{\vec{\theta}}, p_{\vec{\theta}})$. To balance these terms, a scaling factor $\lambda$ is introduced for the regularization loss.  We have treated $\lambda$ as a hyperparameter.

\section{Experimental Settings}
We conducted our experiments using \href{https://github.com/MIND-Lab/OCTIS}{OCTIS} \cite{terragni2021octis}, a publicly available framework to compare and optimize the topic models.

\subsection{Datsets}
We used seven publicly available datasets. Four of them, namely 20Newsgroups (\textbf{20NG}) \cite{twenty_newsgroups_113}, BBCNews (\textbf{BBC}) \cite{greene2006practical}, \textbf{M10}, and \textbf{DBLP} \cite{pan2016triparty}, were obtained in pre-processed form from OCTIS. The remaining three datasets -- SearchSnippets (\textbf{SS}), PascalFlickr (\textbf{Pascal}), and Biomedicine (\textbf{Bio}) -- were sourced from \citep{Qiang2020Short} and pre-processed using OCTIS. Further details about these baselines are provided in Appendix \ref{sec:ap_basline}. Dataset statistics are shown in Table \ref{tab:datasets}.
\begin{table}[ht]
    \centering
    \begin{adjustbox}{width=.75\linewidth}
      \begin{tabular}{lccc} 
        \toprule
        \multicolumn{1}{c}{\textbf{Dataset}} & \textbf{\#Docs} & \textbf{\#Labels} & \textbf{\#Words} \\ \midrule
         \textbf{20NG} & 16309 & 20 & 1612 \\ 
         \textbf{BBC} & 2225 & 5 & 2949  \\
         \textbf{M10} & 8355 & 10 & 1696  \\ 
         \textbf{SS} & 12295 & 8 & 2000  \\ 
         \textbf{Pascal} & 4834 & 20 & 2630  \\ 
         \textbf{Bio}  & 19448 & 20 & 2000 \\
         \textbf{DBLP} & 54595 & 4 & 1513 \\ \bottomrule 
    \end{tabular}
    \end{adjustbox}
    \caption{Statistics of the datasets used. \label{tab:datasets}}
\end{table}
\subsection{Baselines}
We evaluate S2WTM against a range of traditional and neural topic models. Traditional baselines include \textbf{LDA} \cite{blei2003latent}, \textbf{LSI} \cite{dumais2004latent}, and \textbf{NMF} \cite{zhao2017online}, which are well-established for topic extraction.

Neural baselines include \textbf{ETM} \cite{dieng2020topic}, which integrates word embeddings; \textbf{DVAE-TM} and \textbf{DVAE-RSVI-TM} \cite{burkhardt2019decoupling}, based on Dirichlet VAEs; \textbf{ProdLDA} \cite{srivastava2017autoencoding}, which adopts a product-of-experts approach; \textbf{CombinedTM} and \textbf{ZeroshotTM} \cite{bianchi2021cross, bianchi2020pre}, which leverage SBERT embeddings \cite{reimers2019sentence} to incorporate contextualized document representations; \textbf{WTM} \cite{nan2019topic}, based on Wasserstein autoencoders; \textbf{vONT} \cite{xu2023vontss}, a vMF-VAE-based model; and \textbf{ECRTM} \cite{wu2023effective}, which enhances topic diversity through embedding clustering. See Appendix \ref{sec:ap_basline} for baseline details.

\subsection{Automatic Evaluation of Topics}
We assess topic models based on topic quality and distinctiveness. Topic quality, reflecting interpretability and coherence, is measured using NPMI \citep{lau2014machine} and CV \citep{roder2015exploring}, as they strongly correlate with human judgment. We exclude UCI \citep{newman2010automatic} and UMass \citep{mimno2011optimizing} due to their weaker correlation with human assessments \citep{hoyle2021automated}.

Topic diversity assesses the degree to which topics are distinct from each other. To evaluate this, we employ IRBO \citep{bianchi2020pre}, as well as its word embedding-based extension, wI-C (Centroid) \citep{terragni2021word}. Higher NPMI and CV scores indicate better topic coherence, while higher IRBO and wI-C scores denote greater topic diversity.

\begin{table}[ht]
    \centering
    \begin{adjustbox}{width=\linewidth}
        \begin{tabular}{l c c c c c} \toprule
            \textbf{Dataset} & \textbf{$M$} & $p_{\vec{\theta}}$ & $N_{\operatorname{batch}}$ & $p_{\text{drop}}$ & \textbf{$\lambda$} \\ \midrule
            \textbf{20NG} & 4000 & $\vmf$ & 1024 & 0.5 & 8.526 \\
            \textbf{BBC} & 8000 & $\vmf$ & 256 & 0.05 & 5.567 \\
            \textbf{M10} & 2000 & $\mathcal{U}(\mathcal{S}^{K-1})$ & 64 & 0.5 & 7.065 \\
            \textbf{SS} & 1000 & $\mvmf$ & 128 & 0.5 & 3.838 \\
            \textbf{Pascal} & 500 & $\mvmf$ & 64 & 0.5 & 0.879 \\
            \textbf{Bio} & 1000 & $\mathcal{U}(\mathcal{S}^{K-1})$ & 1024 & 0.5 & 3.701 \\
            \textbf{DBLP} & 1000 & $\mathcal{U}(\mathcal{S}^{K-1})$ & 512 & 0.2 & 7.018 \\ \bottomrule
        \end{tabular}
    \end{adjustbox}
    \caption{Hyperparameter values of S2WTM. \label{tab:hyp_params}}
\end{table}

\begin{table*}[!t]
    \centering
    \begin{adjustbox}{width=\linewidth}
    \begin{tabular}{ l cc cc cc cc cc cc cc cc }
    \toprule
     \multirow{2}{*}{\textbf{Model}} & \multicolumn{2}{c|}{\textbf{20NG}} & \multicolumn{2}{c|}{\textbf{BBC}} & \multicolumn{2}{c|}{\textbf{M10}} & \multicolumn{2}{c|}{\textbf{SS}} & \multicolumn{2}{c|}{\textbf{Pascal}} & \multicolumn{2}{c|}{\textbf{Bio}} & \multicolumn{2}{c}{\textbf{DBLP}}\\
     \cmidrule(lr){2-3} \cmidrule(lr){4-5} \cmidrule(lr){6-7} \cmidrule(lr){8-9} \cmidrule(lr){10-11} \cmidrule(lr){12-13} \cmidrule(lr){14-15}
        & \textbf{NPMI} & \textbf{CV} & \textbf{NPMI} & \textbf{CV} & \textbf{NPMI} & \textbf{CV} & \textbf{NPMI} & \textbf{CV} & \textbf{NPMI} & \textbf{CV} & \textbf{NPMI} & \textbf{CV} & \textbf{NPMI} & \textbf{CV} \\ \midrule
      
      \textbf{LDA} & 0.092 & 0.599 & 0.076 & 0.565 & -0.047 & 0.369 & -0.066 & 0.362 & -0.072 & 0.356 & 0.019 & 0.444 & 0.015 & 0.348 \\

      \textbf{LSI} & 0.006 & 0.457 & 0.064 & 0.539 & 0.001 & 0.390 & -0.062 & 0.314 & -0.045 & 0.293 & -0.026 & 0.320 & 0.009 & 0.334 \\

      \textbf{NMF} & 0.118 & 0.648 & 0.065 & 0.555 & 0.050 & 0.448 & \underline{0.019} & 0.462 & -0.042 & 0.378 & 0.100 & 0.537 & \underline{0.016} & 0.354 \\ 
      
      \textbf{ETM} & 0.066 & 0.564 & 0.070 & 0.581 & 0.018 & 0.374 & 0.000 & 0.420 & -0.020 & 0.310 & -0.062 & 0.182 & -0.059 & 0.160 \\

      \textbf{DVAE-TM} & \underline{0.155} & \underline{0.748} & -0.032 & 0.530 & -0.054 & 0.381 & -0.175 & 0.357 & 0.000 & 0.421 & 0.113 & 0.540 & -0.271 & 0.373 \\

      \textbf{DVAE-RSVI-TM} & 0.146 & \textbf{0.750} & -0.051, & 0.523 & -0.052 & 0.412 & -0.192 & 0.411 & -0.019 & 0.422 &	0.100 & 0.537 &	-0.269 & 0.356  \\

      \textbf{ProdLDA} & 0.107 & 0.660 & 0.010 & 0.639 & 0.027 & 0.481 & -0.009 & 0.560 & -0.023 & 0.414 & 0.107 & 0.594 & -0.065 & 0.472 \\

      \textbf{ZeroshotTM} & 0.103 & 0.653 & 0.038 & 0.673 & 0.041 & 0.481 & 0.017 & \underline{0.565} & \underline{0.005} & 0.428 & {0.133} & 0.604 & -0.062 & 0.474\\

      \textbf{CombinedTM} & 0.107 & 0.655 & 0.017 & 0.683 & \underline{0.059} & 0.490 & 0.018 & 0.531 & -0.002 & 0.421 & \underline{0.133} & \underline{0.608} & -0.065 & \underline{0.485}\\
     
      \textbf{WTM} & 0.046 & 0.505 & -0.006 & 0.454 & -0.052 & 0.298 & -0.013 & 0.405 & -0.089 & 0.298 & 0.052 & 0.434 & -0.044 & 0.202 \\

      \textbf{vONT} & 0.045 & 0.505 & -0.001 & 0.468 & -0.053 & 0.301 & -0.015 & 0.407 & -0.090 & 0.302 & 0.052 & 0.442 & -0.043 & 0.204 \\
     
     \textbf{ECRTM} & -0.089 & 0.416 & \underline{0.170} & \underline{0.804} & -0.445 & \textbf{0.516} & -0.333 & 0.423 & -0.414 & \underline{0.510} & -0.421 & {0.510} & -0.248 & 0.377 \\ \midrule
     
     \textbf{S2WTM} & \textbf{0.167} & 0.723 & \textbf{0.252} & \textbf{0.863} & \textbf{0.101} & \underline{0.492} & \textbf{0.146} & \textbf{0.683} & \textbf{0.045} & \textbf{0.572} & \textbf{0.191} & \textbf{0.663} & \textbf{0.133} & \textbf{0.558} \\ \bottomrule
    \end{tabular}
    \end{adjustbox}
    \caption{Median coherence scores over five runs per metric, with the highest values in \textbf{bold} and the second-highest values \underline{underlined}. \label{tab:coherence}}
\end{table*}

\begin{table*}[!t]
    \centering
    \begin{adjustbox}{width=\linewidth}
    \begin{tabular}{ l cc cc cc cc cc cc cc cc }
    \toprule
     \multirow{2}{*}{\textbf{Model}} & \multicolumn{2}{c|}{\textbf{20NG}} & \multicolumn{2}{c|}{\textbf{BBC}} & \multicolumn{2}{c|}{\textbf{M10}} & \multicolumn{2}{c|}{\textbf{SS}} & \multicolumn{2}{c|}{\textbf{Pascal}} & \multicolumn{2}{c|}{\textbf{Bio}} & \multicolumn{2}{c}{\textbf{DBLP}}\\
     \cmidrule(lr){2-3} \cmidrule(lr){4-5} \cmidrule(lr){6-7} \cmidrule(lr){8-9} \cmidrule(lr){10-11} \cmidrule(lr){12-13} \cmidrule(lr){14-15}
    & \textbf{IRBO} & \textbf{wI-C} & \textbf{IRBO} & \textbf{wI-C} & \textbf{IRBO} & \textbf{wI-C} & \textbf{IRBO} & \textbf{wI-C} & \textbf{IRBO} & \textbf{wI-C} & \textbf{IRBO} & \textbf{wI-C} & \textbf{IRBO} & \textbf{wI-C}  \\ \midrule
     
      \textbf{LDA} & 0.970 & 0.845 & 0.968 & 0.844 & 0.949 & 0.838 & 0.960 & 0.842 & 0.902 & 0.824 & 0.976 & 0.844 & 0.856 & 0.833 \\

      \textbf{LSI} & 0.911 & 0.840 & 0.899 & 0.841 & 0.820 & 0.829 & 0.789 & 0.833 & 0.792 & 0.820 & 0.814 & 0.833 & 0.510 & 0.804 \\

      \textbf{NMF} & 0.970 & 0.844 & 0.963 & 0.845 & 0.956 & 0.840 & 0.944 & 0.842 & 0.931 & 0.828 & 0.972 & 0.843 & \underline{0.892} &  0.831 \\ 
      
      \textbf{ETM} & 0.828 & 0.829 & {0.969} & 0.844 & 0.451 & 0.797 & 0.957 & 0.842 & 0.207 & 0.750 & 0.127 & 0.761 & 0.021 &  0.734 \\
      
      \textbf{DVAE-TM} & 0.986 & \underline{0.850} & 0.979 & 0.849 & \textbf{1.000} & 0.840 & \textbf{1.000} & 0.842 & 0.976 & 0.833 & 0.993 & 0.845 & 0.669 & 0.813	\\

      \textbf{DVAE-RSVI-TM} & 0.987 & \underline{0.850} &	0.994 & 0.849 & \textbf{1.000} & 0.840 & \textbf{1.000} & 0.842 & 0.978 & 0.833 & 0.996 & 0.845 & 0.546 & 0.827	\\

      \textbf{ProdLDA} & 0.991 & \underline{0.850} & \textbf{1.000} & 0.848 & 0.997 & \underline{0.842} & \textbf{1.000} & 0.845 & 0.987 & 0.833 & 0.997 & \underline{0.846} & \textbf{1.000} & \underline{0.845} \\

      \textbf{ZeroshotTM} & 0.991 & \underline{0.850} & \textbf{1.000} & 0.849 & \textbf{1.000} & \underline{0.842} & \textbf{1.000} & 0.844 & 0.987 & \underline{0.834} & 0.996 & 0.845 & \textbf{1.000} & \underline{0.845} \\

      \textbf{CombinedTM} & 0.992 & \underline{0.850} & \textbf{1.000} & 0.848 & {0.999} & \underline{0.842} & \textbf{1.000} & \underline{0.845} & 0.987 & 0.833 & 0.993 & \underline{0.846} & \textbf{1.000} & \underline{0.845} \\
     
      \textbf{WTM} & 0.787 & 0.831 & 0.938 & 0.843 & 0.960 & 0.839 & \underline{0.995} & 0.844 & 0.898 & 0.824 & 0.976 & 0.844 & 0.891 & 0.842 \\

      \textbf{vONT} & 0.887 & 0.835 & \underline{0.983} & 0.844 & 0.847 & 0.830 & 0.933 & 0.840 & 0.873 & 0.819 & 0.819 & 0.829 & 0.313 & 0.779 \\
     
     \textbf{ECRTM} & \textbf{0.996} & 0.843 & \textbf{1.000} & \underline{0.850} & \textbf{1.000} & 0.836 & \textbf{1.000} & 0.840 & \textbf{1.000} &  0.814 & \textbf{1.000} & 0.839 & \textbf{1.000} & \underline{0.845} \\ \midrule
     
     \textbf{S2WTM} & \underline{0.994} & \textbf{0.881} & \textbf{1.000} & \textbf{0.854} & \underline{0.999} & \textbf{0.857} & \textbf{1.000} &  \textbf{0.851} & \underline{0.994} & \textbf{0.868} & \underline{0.998} & \textbf{0.859} & \textbf{1.000} & \textbf{0.847} \\ \bottomrule
    \end{tabular}
    \end{adjustbox}
    \caption{Median diversity scores over five runs per metric, with the highest values in \textbf{bold} and the second-highest values \underline{underlined}. \label{tab:diversity}}
\end{table*}

\subsection{Hyperparameter Tuning}
In the S2WTM, hyperparameter tuning was carried out for each dataset using Bayesian Optimization in OCTIS, maximizing the NPMI score. The tuned parameters, listed in Table \ref{tab:hyp_params}, include the number of projections ($M$), prior distribution ($p_{\vec{\theta}}$), batch size ($N_{\operatorname{batch}}$), dropout probability ($p_{\text{drop}}$), and scaling factor ($\lambda$). The number of topics was set to match the number of class labels in the dataset, and training was conducted for 100 epochs.

\section{Results and Discussions}\label{sec:results}
We categorize our findings into the following sections: (1) quantitative evaluation (Section \ref{sec:quantative}), (2) extrinsic evaluation (Section \ref{sec:extrinsic}), (3) qualitative evaluation (Section \ref{sec:qualitative}), and (4) LLM-based topic quality evaluations (Section \ref{sec:LLM_eval}).

\subsection{Quantitative Evaluation} \label{sec:quantative}
In the quantitative evolution, we have evaluated the topic models using automatic topic evaluation scores.

\paragraph{Experimental Setup:} For each dataset, we set the topic count $K$ equal to the number of ground-truth labels, which are available for all datasets used. The values of $K$ for \textbf{20NG}, \textbf{BBC}, \textbf{M10}, \textbf{SS}, \textbf{Pascal}, \textbf{Bio}, and \textbf{DBLP} are 20, 5, 10, 8, 20, 20, and 4, respectively. For the robustness of the results, we have reported the median value over 5 random runs for a given model, a given dataset, and a given topic count.

\begin{table*}[!t]
\centering
\begin{adjustbox}{width=.92\linewidth}
  \begin{tabular}{ l l }
    \toprule
    \textbf{Model} & \textbf{Topics} \\ \midrule     
    \multirow{3}{*}{\textbf{LDA}} 
        & game, team, \textcolor{blue}{year}, player, play, win, \textcolor{blue}{good}, hit, season, run \\
        & card, drive, disk, system, \textcolor{blue}{work}, \textcolor{blue}{problem}, driver, machine, run, memory \\ 
        & key, president, chip, government, \textcolor{blue}{make}, encryption, security, \textcolor{blue}{option}, \textcolor{blue}{press}, phone \\ \midrule
         
    \multirow{3}{*}{\textbf{LSI}}
        & \textcolor{blue}{entry}, team, game, \textcolor{blue}{send}, \textcolor{blue}{graphic}, season, \textcolor{blue}{mail}, play, \textcolor{blue}{year}, \textcolor{blue}{list} \\
        & drive, \textcolor{blue}{atheist}, window, graphic, scsi, bit, administration, program, \textcolor{blue}{master}, \textcolor{blue}{government} \\
        & key, encryption, bit, chip, call, \textcolor{blue}{launch}, secret, block, war, \textcolor{blue}{widget} \\ \midrule
         
    \multirow{3}{*}{\textbf{NMF}}
         & game, team, season, play, \textcolor{blue}{year}, player, win, \textcolor{blue}{good}, score, \textcolor{blue}{draft} \\
         & drive, disk, controller, hard, \textcolor{blue}{bio}, \textcolor{blue}{support}, card, feature, scsi, rom \\
         & internet, privacy, information, secure, computer, security, mail, network, address, \textcolor{blue}{service} \\ \midrule
         
    \multirow{3}{*}{\textbf{ETM}}
        & game, play, win, team, \textcolor{blue}{year}, player, \textcolor{blue}{good}, hit, season, score \\
        & drive, disk, card, system, run, bit, window, scsi, \textcolor{blue}{problem}, monitor \\
        & key, chip, encryption, government, bit, clipper, message, system, algorithm, phone \\ \midrule

     \multirow{3}{*}{\textbf{DVAE-TM}}
       & game, team, playoff, score, season, shot, player, penalty, \textcolor{blue}{make}, hockey \\
       & scsi, controller, ide, disk, boot, drive, \textcolor{blue}{people}, mhz, card, connector \\
       & encryption, key, clipper, chip, escrow, \textcolor{blue}{entry}, enforcement, privacy, encrypt, serial \\ \midrule

    \multirow{3}{*}{\textbf{DVAE-RSVI-TM}}
       & playoff, game, baseball, team, hockey, fan, \textcolor{blue}{pen}, \textcolor{blue}{dog}, score, season \\
       & scsi, controller, ide, disk, \textcolor{blue}{left}, meg, boot, bus, card, mhz \\
       & encryption, key, clipper, chip, \textcolor{blue}{population}, enforcement, secure, serial, encrypt, agency \\ \midrule
        
    \multirow{3}{*}{\textbf{ProdLDA}}
        & game, goal, play, score, \textcolor{blue}{wing}, \textcolor{blue}{ranger}, period, tie, playoff, \textcolor{blue}{blue} \\
        & card, drive, driver, board, \textcolor{blue}{problem}, controller, bus, disk, scsi, port \\
        & chip, phone, make, key, bit, clipper, algorithm, block, conversation, secret \\ \midrule
         
    \multirow{3}{*}{\textbf{ZeroshotTM}}
        & \textcolor{blue}{year}, game, team, \textcolor{blue}{good}, season, play, hit, player, league, \textcolor{blue}{average} \\
        & drive, scsi, disk, card, boot, \textcolor{blue}{problem}, bus, board, transfer, ide \\
        & chip, agency, law, key, algorithm, clipper, secure, security, encryption, secret \\ \midrule

    \multirow{3}{*}{\textbf{CombinedTM}}
        & game, \textcolor{blue}{good}, \textcolor{blue}{year}, team, play, season, player, win, hit, league\\
        & card, drive, driver, scsi, bus, \textcolor{blue}{problem}, board, memory, port, cpu \\
        & chip, key, phone, law, warrant, illegal, clipper, encryption, cop, police \\ \midrule

    \multirow{3}{*}{\textbf{WTM}}
        & game, team, player, play, year, \textcolor{blue}{good}, win, season, \textcolor{blue}{make}, time \\
        & drive, \textcolor{blue}{work}, \textcolor{blue}{price}, \textcolor{blue}{good}, \textcolor{blue}{sell}, \textcolor{blue}{make}, \textcolor{blue}{buy}, monitor, card, \textcolor{blue}{system} \\
        & key, government, {system}, encryption, \textcolor{blue}{make}, {program}, \textcolor{blue}{time}, \textcolor{blue}{space}, chip, {number} \\ \midrule

    \multirow{3}{*}{\textbf{vONT}}
        & game, year, good, time, make, team, car, win, play, player \\
        & drive, \textcolor{blue}{problem}, card, run, system, disk, scsi, driver, bus, window \\
        & key, chip, encryption, government, bit, system, \textcolor{blue}{make}, clipper, \textcolor{blue}{time}, phone \\ \midrule

    \multirow{3}{*}{\textbf{ECRTM}}
       & baseball, fan, pitcher, player, \textcolor{blue}{expansion}, pitch, league, \textcolor{blue}{draft}, \textcolor{blue}{suck}, \textcolor{blue}{apple} \\
       & scsi, card, ide, pin, \textcolor{blue}{ranger}, modem, port, ram, mouse, disk \\
       & illegal, transmit, \textcolor{blue}{patient}, warrant, \textcolor{blue}{taxis}, \textcolor{blue}{fund}, restriction, secret, \textcolor{blue}{budget}, \textcolor{blue}{crack} \\ \midrule 

    \multirow{3}{*}{\textbf{S2WTM}}
        & game, team, win, score, player, playoff, goal, play, stat, season \\
        & drive, card, scsi, ide, bus, controller, driver, disk, system, ram \\
        & encryption, secure, chip, encrypt, phone, secret, communication, clipper, agency, security \\ 
    \bottomrule
  \end{tabular}
\end{adjustbox}
\caption{Three representative topics from the 20NG dataset ($K=20$), with unrelated words highlighted in \textcolor{blue}{blue}.}
\label{tab:topics_20NG}
\end{table*}

\paragraph{Findings:} Table \ref{tab:coherence} presents the coherence scores for all models across the evaluated datasets. S2WTM consistently achieves the highest coherence scores in both NPMI and CV across all datasets, with two exceptions. On the \textbf{M10} dataset, S2WTM ranks second in CV score behind ECRTM. On the \textbf{20NG} dataset, it ranks third in CV score, following DVAE-RSVI-TM and DVAE-TM. However, S2WTM still achieves the highest NPMI score on both M10 and 20NG, indicating strong overall topic coherence. These results highlight the consistent superiority of S2WTM in coherence compared to existing models.

Table \ref{tab:diversity} reports the diversity scores for all models. S2WTM achieves the highest wI-C diversity score across all datasets. Although ECRTM generally performs well in terms of IRBO scores, due to its embedding clustering regularization approach, S2WTM performs comparably on the \textbf{20NG}, \textbf{M10}, \textbf{Pascal}, and \textbf{Bio} datasets, and outperforms ECRTM on the remaining ones.

\begin{figure*}
    \centering
    \includegraphics[width=.85\linewidth]{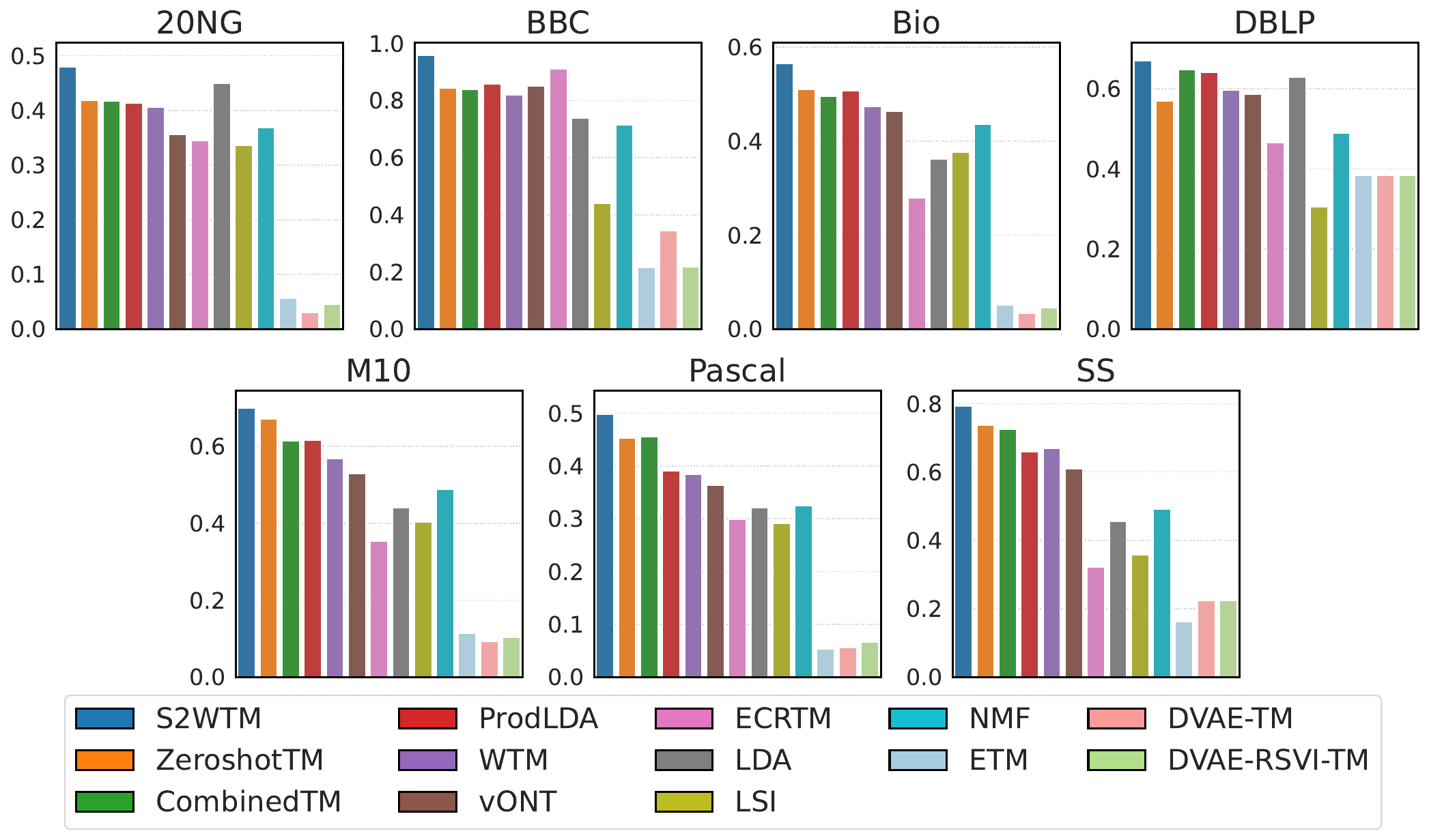}
    \caption{Median document classification accuracy of the models for all datasets. \label{fig:Doc_classification}}
\end{figure*}

\subsection{Qualitative Evaluation} \label{sec:qualitative}
As topic models are unsupervised, their evaluation should go beyond automated coherence scores and include manual inspection \cite{hoyle2021automated}.

\paragraph{Experimental Setup:} We performed a qualitative analysis using the \textbf{20NG} dataset, training all models with 20 topics. Three topics were randomly selected from those generated by S2WTM. To compare these topics across models, we applied the topic alignment algorithm proposed by \citet{adhya2023neural} (see Appendix \ref{sec:ap_alignment_algo}), which identifies the best-matching topics between different models. Table \ref{tab:topics_20NG} presents the aligned topics, ensuring that each row represents semantically similar topics across models. The table also includes NPMI scores for each topic and a match score indicating its similarity to the corresponding topic in S2WTM. Unrelated words in a topic are \textcolor{blue}{highlighted} manually.

\paragraph{Findings:} The three topics in the table broadly represent \textit{sports}, \textit{hardware}, and \textit{encryption/security}. Across all three, S2WTM consistently generates highly coherent and semantically meaningful topics. The \textit{sports} topic includes terms like ``game,'' ``team,'' ``win,'' ``score,'' ``player,'' etc., forming a well-defined theme without unrelated words. The \textit{hardware} topic consists solely of relevant terms such as ``drive,'' ``card,'' ``scsi,'' ``ide,'' ``bus,'' etc., avoiding noise present in other models. Similarly, the \textit{encryption/security} topic includes ``encryption,'' ``secure,'' ``chip,'' ``encrypt,'' ``phone,'' etc., forming a clear, interpretable theme.

\subsection{Extrinsic Evaluation} \label{sec:extrinsic}
To further evaluate the quality of the topic models, we incorporated an extrinsic task focused on document classification. This task assesses how well the learned topic representations can support downstream applications.

\paragraph{Experimental Setup:} We partitioned each of the seven datasets into 70\% train, 15\% validation, and 15\% test. Each topic model was trained using a number of topics equal to the number of labels in the corresponding dataset. The resulting \textit{document-topic} representations (i.e., the topic vector for each document) were then used to train a linear SVM classifier. Fig. \ref{fig:Doc_classification} shows the median accuracy over five runs for all models across all the datasets.

\paragraph{Findings:} As shown in Fig.~\ref{fig:Doc_classification}, S2WTM consistently achieved the highest classification accuracy across all datasets, indicating that it learns more discriminative \textit{document-topic} representations compared to other models. The superior performance of S2WTM is further supported by its higher NMI and Purity scores (detailed in Table \ref{tab:ap_nmi_purity} in the Appendix), suggesting that the learned topic distributions align well with ground truth labels and form coherent clusters. These results collectively demonstrate the effectiveness of S2WTM for both predictive and clustering-based downstream tasks.

\subsection{LLM-based Topic Quality Assessment} \label{sec:LLM_eval}
LLM-based Topic Quality Assessment serves as a proxy for human interpretability, addressing the limitations of automated topic quality evaluations.
\paragraph{Experimental Setup:} Following \citet{stammbach2023revisiting}, we employ two LLM-based evaluation tasks: \textit{rating} and \textit{intrusion detection}. In the rating task, an LLM scores topic coherence from 1 to 3 (``1" = not very related, ``2" = moderately related, ``3" = very related).

In the intrusion detection task, an unrelated word is inserted among the top topic words, and the LLM must identify it. This assumes that intruders are easier to detect in well-defined topics but harder in incoherent ones \cite{chang2009reading}. While LLM-based evaluations have limitations \citep{li2024llms}, they provide valuable insights into topic quality. We used \hyperlink{https://openai.com/index/gpt-4/}{\texttt{GPT-4}} for these evaluations. All models were trained on the \textbf{20NG} dataset with 20 topics.

\begin{figure}
    \centering
    \includegraphics[width=\linewidth]{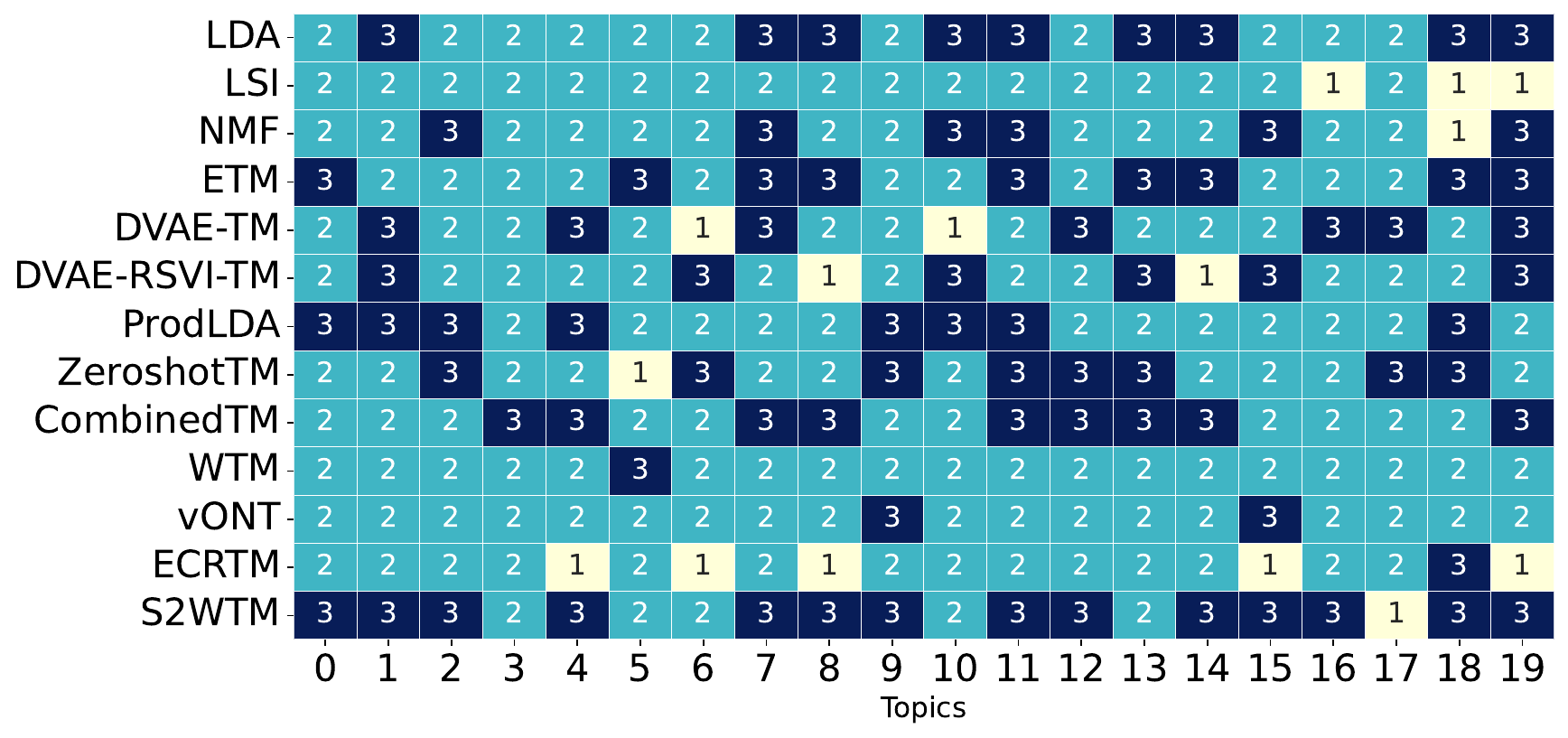}
    \caption{LLM-assigned topic ratings for all models on 20NG. \label{fig:LLM_Rating}}
\end{figure}

\paragraph{Findings in rating task:} Fig. \ref{fig:LLM_Rating} shows the LLM's scores for the rating task, demonstrating that S2WTM generates the highest-quality topics among the evaluated models. S2WTM achieves the highest mean score of 2.7 and a median score of 3.0, indicating consistently high ratings for its topics.

\begin{figure}
    \centering
    \includegraphics[width=\linewidth]{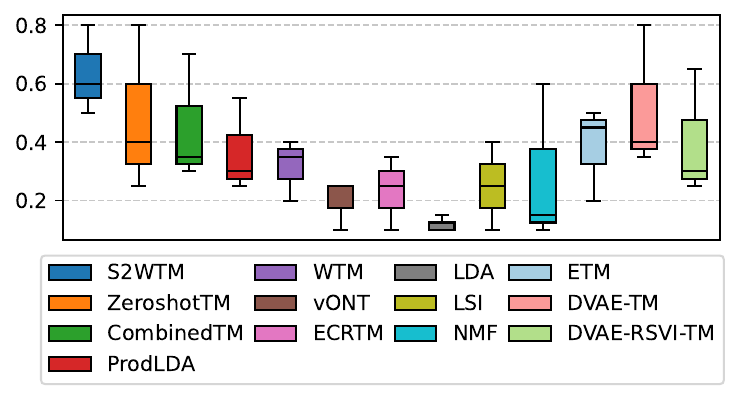}
    \caption{Intruder word detection accuracy over three test runs per model on 20NG. \label{fig:LLM_Intrusion Detection}}
\end{figure}

\paragraph{Findings in intrusion detection task:} We performed three test series for each of the 20 topics generated by each model. For each series, we computed the accuracy of correctly detecting the intruder word. The statistics of these accuracy scores across the three tests are shown in Fig. \ref{fig:LLM_Intrusion Detection}, which again demonstrates the high quality of S2WTM-generated topics compared to existing models.

\section{Beyond Euclidean Assumptions: Empirical Insights}
We aim to evaluate the impact of modeling the latent space on a hypersphere instead of in Euclidean space and its effect on topic quality.

\paragraph{Experimental Setup:} To assess this, we modified the objective function and prior distribution of S2WTM. Specifically, we replaced the Spherical Sliced-Wasserstein distance with the standard Sliced-Wasserstein distance for distribution matching and substituted the hyperspherical prior with a Dirichlet distribution. All other parameters remained unchanged. Table \ref{tab:Euclid_vs_Sph} reports median topic quality scores over five random runs.

\paragraph{Findings:} Comparing the results of S2WTM in Tables \ref{tab:coherence} and \ref{tab:diversity}, NPMI improves by 54.6\% on 20NG and over 100\% on other datasets. CV gains range from 15.9\% to 51.2\%, except for M10, where it remains similar. IRBO and wI-C scores improve by 1.0\% to 52\% and 2.3\% to 5.9\%, respectively, highlighting the benefits of modeling the latent space on a hypersphere.

\begin{table}[!ht]
    \centering
    \begin{adjustbox}{width=\linewidth}
    \begin{tabular}{l cc cc}
        \toprule
        \multirow{2}{*}{\textbf{Dataset}} & \multicolumn{2}{c|}{\textbf{Coherence}} & \multicolumn{2}{c}{\textbf{Diversity}} \\
        \cmidrule(lr){2-3} \cmidrule(lr){4-5}
         & \textbf{NPMI} & \textbf{CV} & \textbf{IRBO} & \textbf{wI-C} \\ \midrule
        \textbf{20NG}    & 0.108 (0.167) & 0.623 (0.723) & 0.984 (0.994) & 0.821 (0.881) \\
        \textbf{BBC}     & 0.081 (0.252) & 0.592 (0.863) & 0.941 (1.000) & 0.835 (0.854) \\
        \textbf{M10}     & 0.038 (0.101) & 0.494 (0.492) & 0.964 (0.999) & 0.826 (0.857) \\
        \textbf{SS}      & 0.024 (0.146) & 0.549 (0.683) & 0.940 (1.000) & 0.809 (0.851) \\
        \textbf{Pascal}  & 0.007 (0.045) & 0.425 (0.572) & 0.887 (0.994) & 0.826 (0.868) \\
        \textbf{Bio}     & 0.094 (0.191) & 0.572 (0.663) & 0.972 (0.998) & 0.818 (0.859) \\
        \textbf{DBLP}    & 0.004 (0.133) & 0.369 (0.558) & 0.658 (1.000) & 0.838 (0.847) \\ \bottomrule
    \end{tabular}
    \end{adjustbox}
    \caption{Median scores over five runs without the spherical latent space assumption. Corresponding S2WTM scores are shown in parentheses. \label{tab:Euclid_vs_Sph}}
\end{table}

\section{Conclusion}
S2WTM is a Spherical Sliced-Wasserstein Autoencoder-based topic model that models the latent space as hyperspherical instead of Euclidean while avoiding posterior collapse seen in VAE-based models. We evaluated S2WTM on seven publicly available datasets for topic modeling assessment. In most experiments, S2WTM consistently achieved superior topic coherence and diversity compared to competitive topic models from the literature. Manual evaluation of selected topics further confirmed that S2WTM produces more coherent topics than alternative models. Additionally, in extrinsic evaluations, S2WTM outperformed existing models across all datasets. Finally, our LLM-based evaluation, serving as a proxy for human judgment, further demonstrated that S2WTM generates higher-quality topics.

\section*{Limitations}
While S2WTM performs well, computing SSW requires multiple spherical Radon transforms, leading to increased computational cost as the number of projections grows.

Additionally, in Section \ref{sec:LLM_eval}, we employ LLMs for topic quality evaluation. However, LLMs are known to exhibit biases \cite{li2024llms, guo2024bias, kotek2023gender}. To mitigate this as much as possible, we conducted a manual evaluation of randomly selected topics (see Section \ref{sec:qualitative}).

\section*{Ethics Statement}
The topic words presented in Table \ref{tab:topics_20NG} are generated by topic models trained on the \textbf{20NG} dataset. These models learn patterns from data without explicit supervision, and their outputs may reflect biases present in the dataset. The authors have no intention to cause harm or offense to any community, religion, country, or individual. Efforts have been made to ensure that the generated topics are analyzed objectively, and any unintended biases in the model outputs are acknowledged as part of the limitations of automated topic modeling.


\appendix

\section{ELBO Surgery and Posterior Collapse} \label{sec:ap_elbo_surgery}
In VAEs, \textit{posterior collapse} refers to the phenomenon where the approximate posterior $q(\mathbf{z}|\mathbf{x})$ becomes indistinguishable from the prior $p(\mathbf{z})$, resulting in latent variables that carry little to no information about the input $\mathbf{x}$. This can be formally characterized as follows:

\begin{definition}[$\epsilon$-Posterior Collapse \cite{kinoshita2023controlling}]\label{def:Eps_PC}
For a given parameter $\theta \in \mathbb{R}^K$, a dataset $\vec{x} \in \mathbb{R}^{m \times n}$, and a closeness criterion $d(\cdot, \cdot)$, the $\epsilon$-posterior collapse is defined for a given $\epsilon \geq 0$, as  
\[
d\left(q_{\vec{\phi}}(z_i|x_i), p(z_i)\right)<\epsilon, \quad \forall i=1,...,n.
\]
\end{definition}

Typically, the closeness criterion $d(\cdot, \cdot)$ is chosen as the Kullback--Leibler divergence $\KL (\cdot \| \cdot)$, consistent with its use in the Evidence Lower Bound (ELBO) objective:

\begin{equation*}
\mathcal{L}_{\text{ELBO}} = \mathbb{E}_{q(\mathbf{z}|\mathbf{x})}[\log p(\mathbf{x}|\mathbf{z})] - \text{KL}(q(\mathbf{z}|\mathbf{x}) \| p(\mathbf{z}))
\end{equation*}

To understand the effect of the KL divergence term, \citet{hoffman2016elbo} introduced a decomposition known as \textit{ELBO surgery}, which analyzes the expected KL term under the data distribution $p(\mathbf{x})$:

\begin{align*}
\mathbb{E}&_{p(\mathbf{x})} \left[ \text{KL}(q(\mathbf{z}|\mathbf{x}) \| p(\mathbf{z})) \right] \\
&= \mathbb{E}_{p(\mathbf{x})} \left[ \int q(\mathbf{z}|\mathbf{x}) \log \frac{q(\mathbf{z}|\mathbf{x})}{p(\mathbf{z})} d\mathbf{z} \right] \\
&= \mathbb{E}_{p(\mathbf{x})} \left[ \int q(\mathbf{z}|\mathbf{x}) \left( \log \frac{q(\mathbf{z}|\mathbf{x})}{q(\mathbf{z})} + \log \frac{q(\mathbf{z})}{p(\mathbf{z})} \right) d\mathbf{z} \right]\\
& = \underbrace{\mathbb{E}_{p(\mathbf{x})} \left[ \text{KL}(q(\mathbf{z}|\mathbf{x}) \| q(\mathbf{z})) \right]}_{I_q(\mathbf{x}; \mathbf{z})} + \underbrace{\text{KL}(q(\mathbf{z}) \| p(\mathbf{z}))}_{\text{Marginal KL}}
\end{align*}

Thus, we obtain the decomposition:

\begin{equation*}
\mathbb{E}_{p(\mathbf{x})} \left[ \text{KL}(q(\mathbf{z}|\mathbf{x}) \| p(\mathbf{z})) \right] = I_q(\mathbf{x}; \mathbf{z}) + \text{KL}(q(\mathbf{z}) \| p(\mathbf{z}))
\end{equation*}

This decomposition reveals that the KL term implicitly includes a \textit{mutual information} component $I_q(\mathbf{x}; \mathbf{z})$, which measures how much information about $\mathbf{x}$ is encoded in the latent variable $\mathbf{z}$. Minimizing the KL term too aggressively can therefore inadvertently reduce mutual information, leading to \textit{posterior collapse}, where the encoder learns to ignore $\mathbf{x}$ and produces non-informative latent representations.

\section{Sampling from the vMF Distribution} \label{sec:ap_prior}  
Sampling from the vMF distribution \cite{ulrich1984computer} involves generating two components:  
\begin{itemize}
    \item A \textit{tangential component} $\vec{v} \sim \mathcal{S}^{d-2}$.
    \item A \textit{radial component} $t$ sampled from $f_{radial}(t|\kappa, p)$.
\end{itemize}

An intermediate sample $\vec{z}' \in \mathbb{R}^p$ is then computed as:  
\begin{equation}
    \vec{z}' = t \vec{e}_1 + \sqrt{1 - t^2} \vec{v}
\end{equation}
where $\vec{e}_1 = (1,0,\dots,0)$ is a unit vector. The final sample is obtained via the Householder transformation:  
\begin{equation}
    \vec{z} = \vec{H} \vec{z}'
\end{equation}
where $\vec{H}$ reflects vectors to align $\vec{e}_1$ with the desired mean direction $\vec{\mu}$.

\section{Datasets Description} \label{sec:ap_dataset}
We used seven publicly available datasets from diverse domains, including news, academia, biomedical literature, and web search, to evaluate topic modeling approaches.

\begin{table*}[!t]
    \centering
    \resizebox{\textwidth}{!}{%
    \begin{tabular}{lcccccc}
    \toprule
    \textbf{Dataset} & \textbf{M=250} & \textbf{M=500} & \textbf{M=1000} & \textbf{M=2000} & \textbf{M=4000} & \textbf{M=8000} \\
    \midrule
    \textbf{20NG} & (0.1456, 5.903) & (0.1453, 6.869) & (0.1519, 7.467) & (0.1518, 10.489) & \highlightpair{(0.1670, 13.515)} & (0.1649, 22.128) \\
    \textbf{BBC} & (0.2139, 3.440) & (0.2219, 5.474) & (0.2501, 7.649) & (0.2371, 9.102) & (0.2437, 11.175) & \highlightpair{(0.2523, 20.239)} \\
    \textbf{M10} & (0.0642, 5.800) & (0.0772, 6.185) & (0.0976, 9.764) & \highlightpair{(0.1010, 11.665)} & (0.0990, 17.658) & (0.1007, 32.467) \\
    \textbf{SS} & (0.1142, 7.322) & (0.1165, 8.132) & \highlightpair{(0.1456, 9.065)} & (0.1455, 11.765) & (0.1450, 18.792) & (0.1445, 33.772) \\
    \textbf{Pascal} & (0.0140, 4.922) & \highlightpair{(0.0446, 6.641)} & (0.0406, 7.601) & (0.0427, 9.326) & (0.0420, 12.755) & (0.0435, 21.093) \\
    \textbf{Bio} & (0.1767, 4.856) & (0.1872, 5.465) & \highlightpair{(0.1905, 7.346)} & (0.1904, 10.642) & (0.1884, 17.374) & (0.1897, 31.194) \\
    \textbf{DBLP} & (0.1257, 10.629) & (0.1261, 11.361) & \highlightpair{(0.1333, 12.074)} & (0.1303, 16.006) & (0.1300, 21.938) & (0.1329, 37.242) \\
    \bottomrule
    \end{tabular}%
    }
    \caption{NPMI and Per-Epoch Training Time (in seconds) for Different Numbers of Projections $M$. The best NPMI values are highlighted.}
    \label{tab:npmi_time_vs_m}
\end{table*}

\begin{enumerate}
    \item \textbf{20NewsGroups (20NG)} is a popularly used text dataset for topic modeling, consisting of documents collected from 20 different online newsgroups. It contains a total of 16,309 pre-processed documents. Each document is assigned a label corresponding to its respective newsgroup.
    
    \item \textbf{BBCNews (BBC)} \cite{greene2006practical} is a collection of 2,225 news articles published by the British Broadcasting Corporation (BBC). Each article is assigned to one of five predefined categories: \textit{tech}, \textit{business}, \textit{entertainment}, \textit{sports}, or \textit{politics}.

    \item \textbf{M10} \cite{greene2006practical} is a subset of the CiteSeer$^X$ digital library, consisting of 8,355 academic documents spanning 10 different research topics.
    
    \item \textbf{SearchSnippets (SS)} \cite{Qiang2020Short} is a dataset constructed from web search transactions, consisting of 12,295 short text documents after pre-processing. These documents are derived from predefined search phrases and are categorized into eight thematic domains: \textit{business}, \textit{computers}, \textit{culture-arts}, \textit{education-science}, \textit{engineering}, \textit{health}, \textit{politics-society}, and \textit{sports}.

    \item \textbf{PascalFlickr (Pascal)} \cite{Qiang2020Short} is a dataset consisting of 4,834 image captions (after pre-processing) sourced from Pascal and Flickr image collections. These captions are categorized into 20 different thematic classes.

    \item \textbf{Biomedicine (Bio)}\footnote{The \textbf{SS}, \textbf{Pascal}, and \textbf{Bio} datasets, along with their corresponding labels, are available at: \url{https://github.com/qiang2100/STTM/tree/master/dataset}} \cite{Qiang2020Short} is derived from biomedical literature made available through the BioASQ challenge. It contains 19,448 documents after pre-processing, each assigned to one of 20 biomedical categories.

    \item \textbf{DBLP} \cite{pan2016triparty} is a bibliographic dataset in computer science, curated by selecting conferences from four research domains: \textit{databases}, \textit{data mining}, \textit{artificial intelligence}, and \textit{computer vision}. The dataset comprises 54,595 documents after pre-processing.
\end{enumerate}

The datasets -- \textbf{20NG}, \textbf{BBC}, \textbf{DBLP} and \textbf{M10} are available in the \href{https://github.com/MIND-Lab/OCTIS}{OCTIS} framework. For the remaining three datasets -- \textbf{SS}, \textbf{Pascal}, and \textbf{Bio} -- we performed additional pre-processing to ensure consistency with the OCTIS format and maintain comparability across all datasets. The details of our pre-processing steps are described in Section \ref{sec:ap_preprocess}.

\subsection{Preprocessing} \label{sec:ap_preprocess}  
Using OCTIS, we convert all text to lowercase, remove punctuation, apply lemmatization, filter out words with fewer than three characters, and discard documents containing fewer than three words.  

\section{Baseline Configurations} \label{sec:ap_basline}
We reproduced all baseline models based on their original papers, using either official implementations or OCTIS. For LDA \citep{blei2003latent}, LSI \citep{dumais2004latent}, NMF \citep{zhao2017online}, ETM \citep{dieng2020topic}, ProdLDA \citep{srivastava2017autoencoding}, ZeroShotTM \citep{bianchi2021cross}, and CombinedTM \citep{bianchi2020pre}, we used the default parameter settings available in OCTIS.

For DVAE-TM, DVAE-RSVI-TM \cite{burkhardt2019decoupling} \footnote{DVAE-TM and DVAE-RSVI-TM: \url{https://github.com/mayanknagda/neural-topic-models}}, WTM \citep{nan2019topic} \footnote{WTM: \url{https://github.com/zll17/Neural_Topic_Models/blob/master/models/WTM.py}}, vONT \citep{xu2023vontss} \footnote{vONT: \url{https://github.com/xuweijieshuai/vONTSS}}, and ECRTM \citep{wu2023effective} \footnote{ECRTM: \url{https://github.com/BobXWu/ECRTM}}. These models were integrated into the OCTIS framework to facilitate a standardized evaluation, ensuring fair and reproducible comparisons with other baseline models.

\section{Computing Infrastructure} \label{sec:ap_Computing_Infrastructure}
Our experiments were conducted on a system with an Intel\textsuperscript{\textregistered} Core\textsuperscript{\textregistered} i7-10700K processor, 32 GB of RAM, an NVIDIA GeForce GTX 1660 SUPER GPU with 6 GB of VRAM, CUDA 12.2, and the Ubuntu 22.04 operating system.

\section{Trade-off Between Performance and Computational Cost in Terms of Number of Projections} \label{sec:ap_tradeoff}
The computational complexity of SSW grows linearly with the number of projections $M$, as noted by \citet{bonet2023spherical}. Their empirical analysis confirms that increasing $M$ proportionally increases runtime. Following this, we investigate the trade-off between performance and computational cost for S2WTM by varying $M$ and measuring its impact on model quality.

Table~\ref{tab:npmi_time_vs_m} reports the median NPMI scores across five random runs, paired with the corresponding per-epoch training times, for different values of $M$. We highlight the best NPMI scores for each dataset.

The results suggest that while increasing $M$ generally improves performance, the gains plateau beyond a certain point. Across most datasets, performance stabilizes around $M=1000$, offering a strong balance between topic quality and computational efficiency. Although theoretically determining the optimal $M$ remains an open research question, our empirical findings suggest that setting $M \approx 1000$ yields favorable results without incurring excessive computational costs. These observations are consistent with the linear scaling behavior reported by \citet{bonet2023spherical}.

\begin{table*}[!t]
    \centering
    \begin{adjustbox}{width=\linewidth}
    \begin{tabular}{l cc cc cc cc cc cc cc}
    \hline
    \multirow{2}{*}{\textbf{Model}} & \multicolumn{2}{c|}{\textbf{20NG}} & \multicolumn{2}{c|}{\textbf{BBC}} & \multicolumn{2}{c|}{\textbf{M10}} & \multicolumn{2}{c|}{\textbf{SS}} & \multicolumn{2}{c|}{\textbf{Pascal}} & \multicolumn{2}{c|}{\textbf{Bio}} & \multicolumn{2}{c}{\textbf{DBLP}} \\
    \cmidrule(lr){2-3} \cmidrule(lr){4-5} \cmidrule(lr){6-7} \cmidrule(lr){8-9} \cmidrule(lr){10-11} \cmidrule(lr){12-13} \cmidrule(lr){14-15}
     & NMI & Purity & NMI & Purity & NMI & Purity & NMI & Purity & NMI & Purity & NMI & Purity & NMI & Purity \\
    \hline
    \textbf{LDA} & \underline{0.425} & \underline{0.461} & 0.617 & 0.757 & 0.157 & 0.376 & 0.219 & 0.469 & 0.303 & 0.313 & 0.255 & 0.371 & 0.099 & 0.533 \\
    \textbf{LSI} & 0.282 & 0.326 & 0.250 & 0.231 & 0.259 & 0.395 & 0.111 & 0.342 & 0.295 & 0.274 & 0.364 & 0.410 & 0.044 & 0.449 \\
    \textbf{NMF} & 0.298 & 0.360 & 0.602 & 0.808 & 0.240 & 0.449 & 0.152 & 0.423 & 0.326 & 0.343 & 0.338 & 0.421 & 0.090 & 0.505 \\
    \textbf{ETM} & 0.026 & 0.091 & 0.022 & 0.270 & 0.017 & 0.157 & 0.007 & 0.228 & 0.059 & 0.106 & 0.015 & 0.064 & 0.231 & 0.383 \\
    \textbf{DVAE-TM} & 0.227 & 0.230 & 0.414 & 0.431 & 0.192 & 0.269 & 0.251 & 0.225 & 0.263 & 0.258 & 0.260 & 0.267 & 0.182 & 0.383 \\
    \textbf{DVAE-RSVI-TM} & 0.238 & 0.237 & 0.308 & 0.374 & 0.164 & 0.254 & 0.269 & 0.225 & 0.270 & 0.264 & 0.267 & 0.262 & 0.107 & 0.383 \\
    \textbf{CombinedTM} & 0.406 & 0.418 & 0.650 & 0.838 & 0.437 & 0.614 & 0.503 & 0.725 & 0.463 & 0.459 & 0.407 & 0.496 & 0.240 & 0.647 \\
    \textbf{ZeroshotTM} & 0.383 & 0.421 & 0.637 & 0.844 & 0.460 & 0.671 & 0.509 & 0.737 & 0.465 & 0.459 & 0.406 & 0.511 & 0.168 & 0.569 \\
    \textbf{ProdLDA} & 0.369 & 0.414 & 0.685 & 0.859 & 0.420 & 0.617 & 0.418 & 0.660 & 0.407 & 0.397 & 0.398 & 0.508 & 0.230 & 0.641 \\
    \textbf{WTM} & 0.370 & 0.366 & 0.718 & 0.853 & 0.340 & 0.543 & 0.431 & 0.678 & 0.401 & 0.375 & 0.347 & 0.473 & 0.188 & 0.598 \\
    \textbf{vONT} & 0.328 & 0.351 & 0.721 & 0.851 & 0.351 & 0.529 & 0.402 & 0.629 & 0.420 & 0.363 & 0.342 & 0.440 & 0.179 & 0.572 \\
    \textbf{ECRTM} & 0.336 & 0.345 & 0.716 & 0.861 & 0.142 & 0.355 & 0.084 & 0.322 & 0.315 & 0.317 & 0.167 & 0.281 & 0.059 & 0.465 \\
    \midrule
    \textbf{S2WTM} & \textbf{0.437} & \textbf{0.469} & \textbf{0.729} & \textbf{0.874} & \textbf{0.464} & \textbf{0.680} & \textbf{0.547} & \textbf{0.749} & \textbf{0.471} & \textbf{0.521} & \textbf{0.557} & \textbf{0.644} & \textbf{0.254} & \textbf{0.686} \\
    \hline
    \end{tabular}
    \end{adjustbox}
    \caption{Clustering performance (NMI and Purity) of the models across all the datasets.} \label{tab:ap_nmi_purity}
\end{table*}

\section{Clustering Alignment Metrics: NMI and Purity} \label{sec:ap_nmi_purity}
To complement the document classification results reported in Section~\ref{sec:extrinsic}, we evaluate how well the learned topic distributions align with ground-truth class labels using two clustering metrics: Normalized Mutual Information (NMI) and Purity. While classification accuracy reflects predictive utility, these unsupervised metrics assess the structure and separability of the learned representations.

As shown in Table~\ref{tab:ap_nmi_purity}, S2WTM consistently achieves the highest scores across all datasets. This indicates that the learned topic distributions align well with the true labels and form coherent, well-separated clusters, further confirming the strength of S2WTM for clustering-based downstream tasks.

\section{Topic Alignment Algorithm} \label{sec:ap_alignment_algo}
We align topics between two given models using the following two-step strategy as prescribed by \cite{adhya2023neural}:

\begin{enumerate}
    \item \textbf{Construct Similarity Matrix:}
    \begin{enumerate}
        \item Let $P = \{P[1], P[2], \dots, P[K]\}$ and $Q = \{Q[1], Q[2], \dots, Q[K]\}$ be the topic lists from the two models.
        \item Compute the Rank-biased Overlap (RBO) similarity \cite{webber2010similarity} for each topic pair:
        \begin{align*}
            a_{i,j} &= \operatorname{RBO}\big(P[i], Q[j]\big), \\
            &\quad \forall i,j \in \{1, \dots, K\}.
        \end{align*}
        \item Construct the similarity matrix $\mathbf{A} = (a_{ij})_{1 \leq i,j \leq K}$, where each entry $a_{ij} \in [0,1]$, with $0$ indicating no overlap and $1$ indicating exact overlap.
    \end{enumerate}
    
    \item \textbf{Iterative Topic Pairing:}
    \begin{enumerate}
        \item While there are unaligned topics in $P$ and $Q$:
        \begin{enumerate}
            \item Identify the pair $(i^*, j^*)$ with the highest similarity score:
            \[
            (i^*, j^*) = \arg\max_{i,j} a_{ij}
            \]
            \item Align the topics $P[i^*]$ and $Q[j^*]$.
            \item Exclude the selected topics from further consideration:
            \[
            P \leftarrow P \setminus \{P[i^*]\}, \quad Q \leftarrow Q \setminus \{Q[j^*]\}
            \]
            ensuring that each topic is aligned only once.
        \end{enumerate}
    \end{enumerate}
\end{enumerate}

\end{document}